\documentclass[journal]{IEEEtran}
\usepackage{amsmath}
\usepackage{algorithm}
\usepackage[noend]{algpseudocode}
\usepackage{graphicx}
\usepackage{array}
\usepackage{capt-of}
\graphicspath{ {/} }

\makeatletter
\def\BState{\State\hskip-\ALG@thistlm}
\makeatother
\ifCLASSINFOpdf
\else
\fi
\begin{document}

%
\title{A parallel sampling based clustering}
%
%
%

\author{AV Aditya Sastry,B.Tech, GITAM University
             Kalyan Netti, Senior Scientist, CSIR-National Geophysical Research Institute}
\maketitle

\begin{abstract}
The problem of automatically clustering data is an age old problem. People have created numerous algorithms to tackle this problem. The execution time of any of this algorithm grows with the number of input points and the number of cluster centers required. To reduce the number of input points we could average the points locally and use the means or the local centers as the input for clustering. However since the required number of local centers is very high, running the clustering algorithm on the entire dataset to obtain these representational points is very time consuming. To remedy this problem, in this paper we are proposing two subclustering schemes where by we subdivide the dataset into smaller sets and run the clustering algorithm on the smaller datasets to obtain the required number of datapoints to run our clustering algorithm with. As we are subdividing the given dataset, we could run clustering algorithm on each smaller piece of the dataset in parallel. We found that both parallel and serial execution of this method to be much faster than the original clustering algorithm and error in running the clustering algorithm on a reduced set to be very less.
\end{abstract}

\begin{IEEEkeywords}
Kmeans, clustering algorithm, CUDA, parallel computing.
\end{IEEEkeywords}

%
\IEEEpeerreviewmaketitle

\section{Introduction}
%
%
%
%
\IEEEPARstart {T}he amount of data that we are generating each day is way too huge to be manually categorized by any team, no matter how large it maybe. An automated system which can classify all kinds of data, which scales well to the size of the input is the need of the hour, if we wish to completely leverage the information available to us.\\    

	Various clustering algorithms were examined in Xu et al[1] The monothetic clustering algorithm or MONA proposed by Chavent et al[2] which is a divisive clustering algorithm. The possibilistic fuzzy C-Means clustering by Pal et al[3] uses the fuzzy approach to compute membership of each data point. However in this paper we will be using the traditional approach to clustering, the Lloyd’s algorithm for deriving the local centers in the sub clusters cause of its relative fastness.\\ 

Several papers have explored the idea of subclustering in the past. Karypis et al[4] considers each datapoint as a subcluster initially and merges two subclusters if the proximity between two subclusters is higher than the internal interconnectivity of the points in a subcluster. Savaresi et al[5] examimes the two famous divisive bisecting K means algorithms. Essentially both of them take a larger cluster and divide it into two smaller clusters and continue to do the same on the smaller clusters till the required number of clusters is acquired. These two algorithms provide highly accurate and pure sub clusters but either they involve computing a expensive inter cluster proximity or computing a expensive eigenvalue decomposition. \\        

	In this paper we propose two methods of sub dividing the dataspace into multiple regions. We run clustering algorithm in each of these regions to obtain local cluster centers on which we run clustering algorithm again to obtain the required number of centers. By running clustering algorithm on each region individually we are effectively negating the effect of any outliers present in the region. We found that by using sufficiently large datasets and retrieving sufficiently a lot of cluster centers from a region, the error in clustering is very less and the performance gain is very high.\\    

	In section II of the paper we cover equal sized subclustering algorithm and in section III we cover unequal sized subclustering algorithm. In section IV we cover the parallel implementation of this algorithm and in section V we show experimental results.

\section{Equal sized subclustering} 

	Checking pair wise for similarity between points while sub grouping points is a very time consuming option. Instead we select few points that we will refer to as landmark points and based on each points similarity to a particular landmark we will gather the points into a subgroup. The similarity measure could be a distance measure like Euclidean distance, Manhattan distance or anything. \\	

	A recursive algorithm for doing this would involve choosing a point farthest from all points as the landmark point and gathering all the points similar to this into one subgroup and redoing this process from the left out points. The iterative version of the algorithm is given below. \\  
\begin{algorithm}
\caption{Equal Sized Subclustering}\label{EQUAL SIZED SUBCLUSTERING}
\begin{algorithmic}[1]
\Procedure{Subcluster}{}
\State $ \text{Perform feature scaling on all the attributes. } $
\State $ \text{Make a new point L with each attribute having the } $
\Statex  $ \ \ \ \text{ lowest value among all the points for that attribute.} $
\State $ \text {Gather N points closest to L.}$
\Statex $\ \ \ \   \text{Where N is the number of points in a subcluster.} $
\State $\text{Perform clustering on the N points to}$
\Statex $\ \ \ \text{ obtain required number of points.}$
\State $\text{Remove the N points from the dataset.}$

\EndProcedure{\textbf{End}}
\end{algorithmic}
\end{algorithm}

\begin{center}
\includegraphics[width=2.5in]{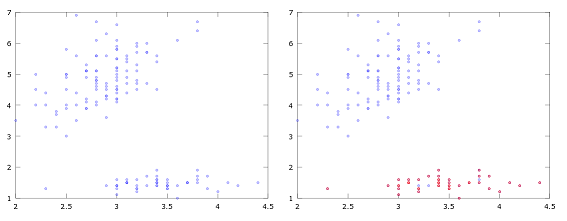}  
\captionof{figure}{The left hand side of the figure shows the original dataset which is a scatter plot between second and third dimmensions of iris dataset. The image on the right shows the same dataset subclustered.}
\end{center}

\section{UNEQUAL SIZED SUBCLUSTERING}

	The problem with the above method is with case where the dataset has way too many outliers. This leads to some of the subclusters being filled only by the outlier points. To prevent that we ideally want the landmarks to be located in the dense regions of the dataset. To ensure that we take a point whose attributes are the lowest values for that attribute throughout our dataset and then we take a point whose attributes are the highest values for that attribute throughout our dataset. Then we connect these two points with a line, subdivide the line into required number of landmarks. Based on the similarity of each datum to these landmarks, they will be subclustered. 

\begin{algorithm}
\caption{Unequal sized subclustering}\label{EQUAL SIZED SUBCLUSTERING}
\begin{algorithmic}[1]
\Procedure{Subcluster}{}
\State $ \text{Perform feature scaling on all the attributes. } $
\State $ \text{Make a new point L with each attribute having }$
\Statex $\ \ \ \  \text{the lowest value among all the points for that} $
\Statex $\ \ \ \   \text{attribute.} $
\State $ \text {Make a new point H with each attribute having }$
\Statex $\ \ \ \   \text{the highest value among all the points for that } $
\Statex $\ \ \ \   \text{attribute} $
\State $\text{Divide the line segment between H and L into }$
\Statex $\ \ \ \ \text{required number of points which are called Landmark}$
\Statex $\ \ \ \ \text{points. }$
\State $\text{Group points according to the landmark point }$
\Statex $\ \ \ \ \text{they are closest to and run clustering algorithms on }$
\Statex $\ \ \ \text{these groups.}$

\EndProcedure{\textbf{End}}
\end{algorithmic}
\end{algorithm}

\begin{center}
\includegraphics[width=2.5in]{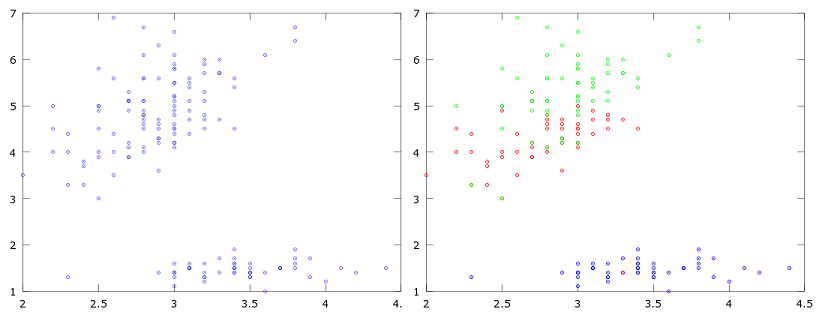}  
\captionof{figure}{The left hand side of the figure shows the original dataset which is a scatter plot between second and third dimmensions of iris dataset. The image on the right shows the same dataset subclustered using the unequal subclustering scheme.}
\end{center}  

\section{CUDA}
	CUDA stands for Compute Unified Device Architecture. CUDA is a parallel computing platform and programming model invented by NVIDIA. It enables dramatic increases in computing performance by harnessing the power of the graphics processing unit (GPU). \\ 

	Owing to the rapid performance growth in Graphic Cards, their ability to perform the same operations concurrently on multiple pieces of data extremely fast and the recent improvements in their programmability GPUs became very lucrative devices for wide range of application domains which involve a lot of data. A lot of research has been presented in recent times to program GPUs for the purposes of general computing.  \\

	There are two parts to a CUDA application. The device part which runs on the GPU is called a kernel. Kernels are implemented in CUDA programming language which is C extended with a bunch of keywords. A kernel is run by multiple threads. A collection of threads is known as a block. Threads within a block can communicate with each other using shared memory. But threads not belonging to the same block cannot communicate with each other. A kernel is called from the Host part of the application by specifying the number of blocks and number of threads in the blocks required. The host part of the application runs on the CPU is responsible for passing the required data to the CUDA kernel and also starting the CUDA kernels.

\section{PARALLEL IMPLEMENTATION}

	As we have mentioned in the previous section, each CUDA application has two parts: the host part and the device part. In the host part we subdivide the given dataset into required number of subclusters using either of our proposed algorithms. The device part of the code is the clustering algorithm with each thread running on one of the subclusters of the original dataset. Once each thread finishes executing and returns the sampled points, we return back to the host part of the application where we run the clustering algorithm on the sampled points to derive the required number of cluster centers.\\ 

	The input to our application is a MxN 2 dimensional array. Where M represents the number of data points and N represents the number of attributes. First we run this 2 dimensional array through either of the sub clustering algorithm to obtain multiple 2 Dimensional arrays representing points in the sub region. Before calling the device code we need to transfer these multiple 2 Dimensional arrays to Device for the kernel to execute on. Converting these multiple 2 Dimensional arrays to 1 Dimensional arrays could take a lot of time. So instead we are generating a 1 dimensional array while subgrouping the points itself. This could be done in the following two ways.\\ 

1)	Row major flattening: we take a given datum and place all of its attributes in consecutive memory locations.\\ 

2)	Column major flattening: we take all values of all datums for a particular attribute place it into consecutive memory locations and then move on to the next attribute.\\ 

After receiving the data from the host the GPU needs to convert the 1 dimensional array back to a 2 dimensional array. This is again done in two ways:\\ 

1)	Row major reconstruction: we take N consecutive values present starting from a given location and turn that into a datum. Where N is the number of attributes.\\ 

2)	Column major reconstruction: We read a first value from a given location and then skip the next N memory locations where N is the number of records. We continue to do this till we obtain M values where M is the number of attributes and construct the required datum.\\ 
	
	We run each subcluster on a block of threads. After all the blocks have run we obtained the data from each block using reconstruction techniques mentioned above and run clustering algorithm on them to obtain the required number of cluster centers.\\

\section{Experimental Results}
	We ran this algorithm on Iris[7] and Seeds[8] dataset for accuracy comparision.

\begin{table}[h]
\begin{tabular}{|c|c|c|}
\hline
                                                                              & Iris                                                                             & Seeds                                                                             \\ \hline
Standard Kmeans                                                               & 133                                                                              & 187                                                                               \\ \hline
\begin{tabular}[c]{@{}c@{}}Equal Partitioning\\ Best performance\end{tabular} & \begin{tabular}[c]{@{}c@{}}138(6 Subclusters\\ 6 times compression)\end{tabular} & \begin{tabular}[c]{@{}c@{}}191(6 Subclusters \\ 6 times compression)\end{tabular} \\ \hline
Unequal Partitioning                                                          & \begin{tabular}[c]{@{}c@{}}138(6 subcluster\\ 6 times compression)\end{tabular}  & \begin{tabular}[c]{@{}c@{}}191(6 Subclusters \\ 6 times compression)\end{tabular} \\ \hline
\end{tabular}
\end{table}

	The Iris dataset has 150 datapoints with 3 classes and 4 attributes. The seeds dataset has 210 datapoints with 3 classes and 7 attributes.\\ 

	We ran these algorithms on a NVIDIA TESLA C2075 GPU. On a system with Intel Xeon E5-1410 cpu running at 3.2 GHZ and 48GB of RAM. Our test dataset was a 2 dimensional synthetic dataset consisting of 100k, 250k, 500k elements. Each of these synthetic dataset contained 500 points per cluster. We used compression values of 5, 10, 15 the following table shows the execution time comparison\\ 

\begin{table}[h]
\begin{tabular}{|l|l|l|}
\hline
Dataset Size & Traditional Kmeans & Parallel Kmeans \\ \hline
100,000      & 2.328              & 2.78            \\ \hline
250,000      & 25.6               & 4.96            \\ \hline
500,000      & 156.8              & 6.2                \\ \hline

\end{tabular}
\end{table}

	Roughly 30 times speedup has been observed for 500k elements with compression value 5 and about 55 times speedup for compression value 20

\begin{table}[h]
\begin{tabular}{|l|l|}
\hline
Compression Value & Execution Time \\ \hline
5                 & 6.2            \\ \hline
10                & 5.76           \\ \hline
15                & 4.83           \\ \hline
20                &                \\ \hline
\end{tabular}
\end{table}


%


%

%

\end{document}